\documentclass[a4paper, 10 pt, conference]{ieeeconf}  % Comment this line out if you need a4paper
\IEEEoverridecommandlockouts                              % This command is only needed if 
\usepackage[utf8]{inputenc}
\usepackage{graphicx} % for pdf, bitmapped graphics files
\usepackage{amsmath} % assumes amsmath package installed
\usepackage{amssymb}  % assumes amsmath package installed
\usepackage{mathtools}
\usepackage{accents}
\usepackage{booktabs}
\usepackage{multirow}
\usepackage{subfig}
\usepackage{xcolor}
\usepackage[hidelinks]{hyperref}
\usepackage{siunitx}
\usepackage[font=small]{caption}
\DeclareMathOperator*{\argmax}{argmax} 
\newcommand{\etal}{\textit{et al}.\@ }
\definecolor{tu0}{rgb}{0.7451, 0.1176, 0.2353}

\definecolor{tu1}{rgb}{1.0000, 0.8039, 0.0000}
\definecolor{tu11}{rgb}{1.0000, 0.8627, 0.3020}
\definecolor{tu12}{rgb}{1.0000, 0.9020, 0.4980}
\definecolor{tu13}{rgb}{1.0000, 0.9412, 0.6980}
\definecolor{tu14}{rgb}{1.0000, 0.9608, 0.8000}

\definecolor{tu2}{rgb}{0.9804, 0.4314, 0.0000}
\definecolor{tu21}{rgb}{0.9882, 0.6039, 0.3020}
\definecolor{tu22}{rgb}{0.9882, 0.7137, 0.4980}
\definecolor{tu23}{rgb}{0.9922, 0.8275, 0.6980}
\definecolor{tu24}{rgb}{0.9961, 0.8863, 0.8000}

\definecolor{tu3}{rgb}{0.6902, 0.0000, 0.2745}
\definecolor{tu31}{rgb}{0.7529, 0.2000, 0.4196}
\definecolor{tu32}{rgb}{0.8431, 0.4980, 0.6353}
\definecolor{tu33}{rgb}{0.9216, 0.7490, 0.8196}
\definecolor{tu34}{rgb}{0.9529, 0.8510, 0.8902}

\definecolor{tu4}{rgb}{0.4863, 0.8039, 0.9020}
\definecolor{tu41}{rgb}{0.6431, 0.8627, 0.9333}
\definecolor{tu42}{rgb}{0.7412, 0.9020, 0.9490}
\definecolor{tu43}{rgb}{0.8431, 0.9412, 0.9686}
\definecolor{tu44}{rgb}{0.8980, 0.9608, 0.9804}

\definecolor{tu5}{rgb}{0.0000, 0.5020, 0.7059}
\definecolor{tu51}{rgb}{0.3020, 0.6510, 0.7961}
\definecolor{tu52}{rgb}{0.5490, 0.7765, 0.8667}
\definecolor{tu53}{rgb}{0.7490, 0.8745, 0.9255}
\definecolor{tu54}{rgb}{0.8510, 0.9255, 0.9569}

\definecolor{tu6}{rgb}{0.0000, 0.3255, 0.4549}
\definecolor{tu61}{rgb}{0.2510, 0.4941, 0.5922}
\definecolor{tu62}{rgb}{0.5490, 0.6941, 0.7529}
\definecolor{tu63}{rgb}{0.7490, 0.8314, 0.8627}
\definecolor{tu64}{rgb}{0.8510, 0.8980, 0.9176}

\definecolor{tu7}{rgb}{0.7765, 0.9333, 0.0000}
\definecolor{tu71}{rgb}{0.8431, 0.9529, 0.3020}
\definecolor{tu72}{rgb}{0.8863, 0.9647, 0.4980}
\definecolor{tu73}{rgb}{0.9333, 0.9804, 0.6980}
\definecolor{tu74}{rgb}{0.9569, 0.9882, 0.8000}

\definecolor{tu8}{rgb}{0.5373, 0.6431, 0.0000}
\definecolor{tu81}{rgb}{0.6784, 0.7490, 0.3020}
\definecolor{tu82}{rgb}{0.7686, 0.8196, 0.4980}
\definecolor{tu83}{rgb}{0.8588, 0.8941, 0.6980}
\definecolor{tu84}{rgb}{0.9059, 0.9294, 0.8000}

\definecolor{tu9}{rgb}{0.0000, 0.4431, 0.3373}
\definecolor{tu91}{rgb}{0.3020, 0.6118, 0.5373}
\definecolor{tu92}{rgb}{0.5490, 0.7490, 0.7020}
\definecolor{tu93}{rgb}{0.7490, 0.8588, 0.8353}
\definecolor{tu94}{rgb}{0.8549, 0.9176, 0.9059}

\definecolor{tu10}{rgb}{0.8000, 0.0000, 0.6000}
\definecolor{tu101}{rgb}{0.8706, 0.3490, 0.7412}
\definecolor{tu102}{rgb}{0.9216, 0.6000, 0.8392}
\definecolor{tu103}{rgb}{0.9608, 0.8000, 0.9216}
\definecolor{tu104}{rgb}{0.9804, 0.8980, 0.9608}

\definecolor{tu110}{rgb}{0.4627, 0.0000, 0.4627}
\definecolor{tu111}{rgb}{0.5961, 0.2510, 0.5961}
\definecolor{tu112}{rgb}{0.7294, 0.4980, 0.7294}
\definecolor{tu113}{rgb}{0.8392, 0.6980, 0.8392}
\definecolor{tu114}{rgb}{0.9216, 0.8510, 0.9216}

\definecolor{tu120}{rgb}{0.4627, 0.0000, 0.3294}
\definecolor{tu121}{rgb}{0.6118, 0.3020, 0.5333}
\definecolor{tu122}{rgb}{0.7569, 0.5490, 0.6980}
\definecolor{tu123}{rgb}{0.8667, 0.7490, 0.8314}
\definecolor{tu124}{rgb}{0.9216, 0.8510, 0.9020}

\definecolor{tu130}{rgb}{0.0314, 0.0314, 0.0314}
\definecolor{tu131}{rgb}{0.3725, 0.3725, 0.3725}
\definecolor{tu132}{rgb}{0.5882, 0.5882, 0.5882}
\definecolor{tu133}{rgb}{0.7529, 0.7529, 0.7529}
\definecolor{tu134}{rgb}{0.8667, 0.8667, 0.8667}

\definecolor{tu140}{rgb}{0.0000, 0.6875, 0.3125}
\title{\LARGE \bf Class-Incremental Learning for Semantic Segmentation\\ Re-Using Neither Old Data Nor Old Labels}

\author{Marvin Klingner$^{*}$, Andreas Bär$^{*}$, Philipp Donn$^{*}$ and Tim Fingscheidt$^{*}$%
\thanks{$^{*}$Marvin Klingner, Andreas Bär, Philipp Donn and Tim Fingscheidt are with the Institute for Communications Technology,
	Technische Universität Braunschweig, Schleinitzstr. 22, 38106 Braunschweig, Germany
	{\tt\small \{m.klingner, andreas.baer, p.donn, t.fingscheidt\}@tu-bs.de}}%
}

\begin{document}

\maketitle
\thispagestyle{empty}
\pagestyle{empty}

\begin{abstract}
While neural networks trained for semantic segmentation are essential for perception in autonomous driving, most current algorithms assume a fixed number of classes, presenting a major limitation when developing new autonomous driving systems with the need of additional classes. In this paper we present a technique implementing class-incremental learning for semantic segmentation without using the labeled data the model was initially trained on. Previous approaches still either rely on labels for both old and new classes, or fail to properly distinguish between them. We show how to overcome these problems with a novel class-incremental learning technique, which nonetheless requires labels only for the new classes. Specifically, (i) we introduce a new loss function that neither relies on old data nor on old labels, (ii) we show how new classes can be integrated in a modular fashion into pretrained semantic segmentation models, and finally (iii) we re-implement previous approaches in a unified setting to compare them to ours. We evaluate our method on the Cityscapes dataset, where we exceed the mIoU performance of all baselines by $3.5\%$ absolute reaching a result, which is only $2.2\%$ absolute below the upper performance limit of single-stage training, relying on all data and labels simultaneously. 
\end{abstract}

\section{Introduction}

For applications in autonomous driving it is essential to have a valid environment model, which for camera-based systems relies on semantic segmentation assigning a class label to each pixel. A major performance gain was achieved by employing fully convolutional neural networks for this task \cite{Long2015}. Henceforth, major improvements in network architecture and training procedure \cite{Ronneberger2015, Zhao2016a, Zhao2018b, Chen2018a, RotaBulo2018, Zhuang2018} boosted the performance of such models without changing the basic concept.\par

\begin{figure}[t]
	\centering	
	\includegraphics[width=1.0\linewidth]{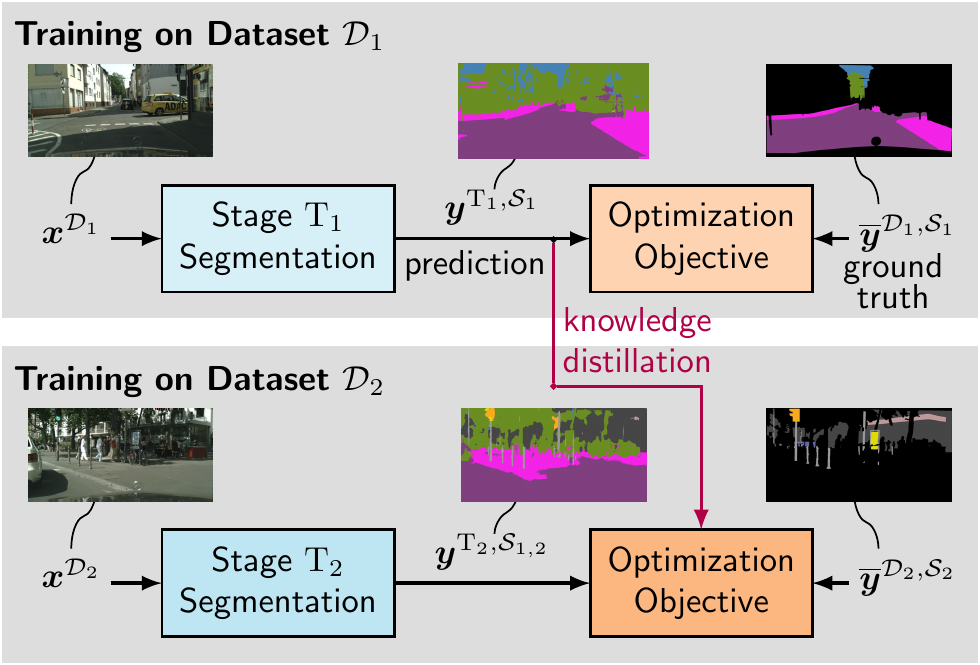}
	\caption{\textbf{General concept} of our class-incremental learning technique for semantic segmentation. Initially, the segmentation model is trained in stage $\mathrm{T}_1$ on dataset $\mathcal{D}_1$ for a set of classes $\mathcal{S}_1$. Afterwards, we extend the model \textbf{in stage $\mathrm{T}_2$} \textbf{by additional classes $\mathcal{S}_2$}, using a second dataset $\mathcal{D}_2$, such that the model outputs both old and new classes $\mathcal{S}_{1,2} = \mathcal{S}_1 \cup \mathcal{S}_2$.}
	\label{fig:general_overview}
\end{figure} 

While current research is typically focused on performance improvements \cite{Chen2018a, RotaBulo2018, Zhuang2018, Zhu2019} or knowledge transfer to new domains \cite{Zou2018, Li2019b, Bolte2019a}, the number of segmented classes is usually assumed to be constant. This, however, presents a major problem when developing new driver assistance systems or autonomous driving capabilities, as these systems may require knowledge about additional classes. A recent approach from image classification \cite{Li2018c} introduced the concept of using the output of a pre-trained teacher model to preserve the knowledge about the old classes, while training the new classes in supervised fashion on new data, labeled only for the new classes. As this approach is \textit{task-incremental}, meaning that the cross-task decision between old and new classes is neither enforced nor preserved, subsequent approaches introduced a memory for further training \cite{Lee2019b}, containing pairs of images and labels from the initially used dataset. This is not only inelegant, but may result in legal data protection issues and may exceed storage capacity, when passing on the labeled data in addition to a pretrained model for class extension.\par
\begin{table*}[t]
  \centering
  \normalsize
  \caption{Overview of employed \textbf{Cityscapes data subsets}: The official training set is splitted into three training subsets. The official Cityscapes validation set serves as test set in this work. In each column the respective (sub)set is defined by the cities of the Cityscapes dataset and the labeled classes we use during training and evaluation of our models, if not mentioned otherwise. Note that class sets $\mathcal{S}_1$, $\mathcal{S}_2$, and $\mathcal{S}_3$ are limited intentionally for experimental investigations and to show the effect of incremental learning.}  
  \setlength{\tabcolsep}{2.2pt}
  \begin{tabular}{l|c|c|c|c}
  	& Training Subset $\mathcal{D}_1$ & Training Subset $\mathcal{D}_2$ & Training Subset $\mathcal{D}_3$ & Our Test Set\\
  	\hline
  	& \multicolumn{3}{c|}{Cityscapes training set} & Cityscapes validation set\\
  	\hline
  	 &&&&\\[-0.25cm]
    Cities & $\left\lbrace\substack{\text{Aachen, Düsseldorf,}\\ \text{Hannover, Strasbourg}}\right\rbrace$ & $\left\lbrace\substack{\text{Bochum, Hamburg,}\\ \text{Jena, Mönchengladbach,}\\ \text{Ulm, Weimar, Tübingen}}\right\rbrace$ & $\left\lbrace\substack{\text{Bremen, Cologne, Stuttgart,}\\ \text{Darmstadt, Krefeld, Zürich}}\right\rbrace$ & $\left\lbrace\substack{\text{Frankfurt, Lindau, Münster}}\right\rbrace$\\
    &&&&\\[-0.25cm]
    \hline
  	&&&&\\[-0.25cm]
    \multirow{3}{*}{\fontsize{14}{14}\selectfont $\substack{\phantom{hhh} \\\phantom{hhh} \\ \text{Labeled}\\ \text{Classes}\;\!}$} & $\mathcal{S}_1\! =\! \left\lbrace\substack{\text{road, sidewalk,}\\ \text{sky, terrain, vegetation}}\right\rbrace$ & $\mathcal{S}_2\! = \!\left\lbrace\substack{\text{building, fence, traffic sign,}\\ \text{pole, traffic light, wall}}\right\rbrace$ & $\mathcal{S}_3\! = \!\left\lbrace\substack{\text{bicycle, car, bus, rider, train,}\\ \text{motorcycle, person, truck}}\right\rbrace$ & $\mathcal{S}_{1,2,3} = \mathcal{S}_1 \cup \mathcal{S}_2 \cup \mathcal{S}_3$\\
    &&&&\\[-0.25cm]
    & \includegraphics[width=0.13\linewidth]{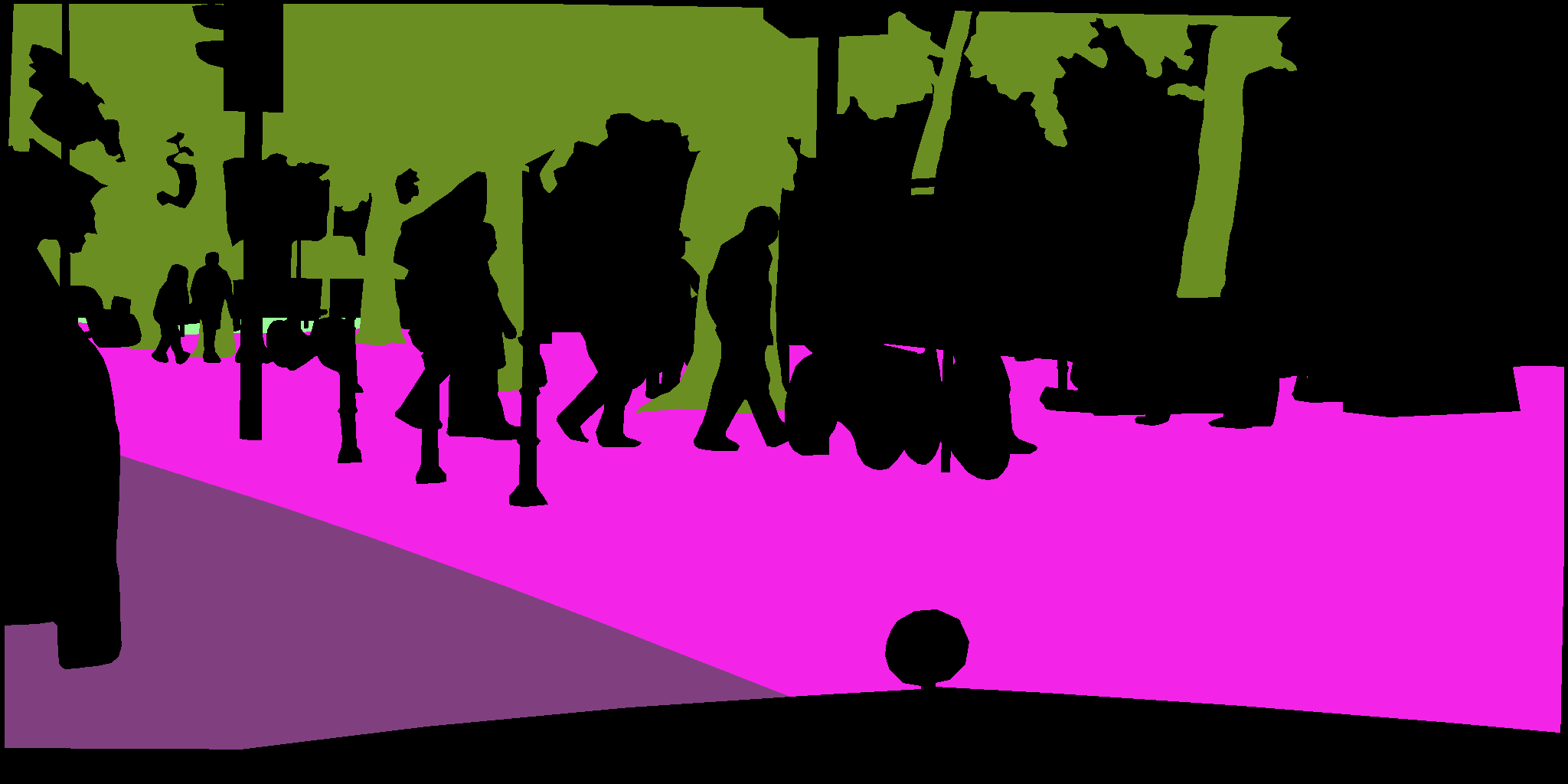} & \includegraphics[width=0.13\linewidth]{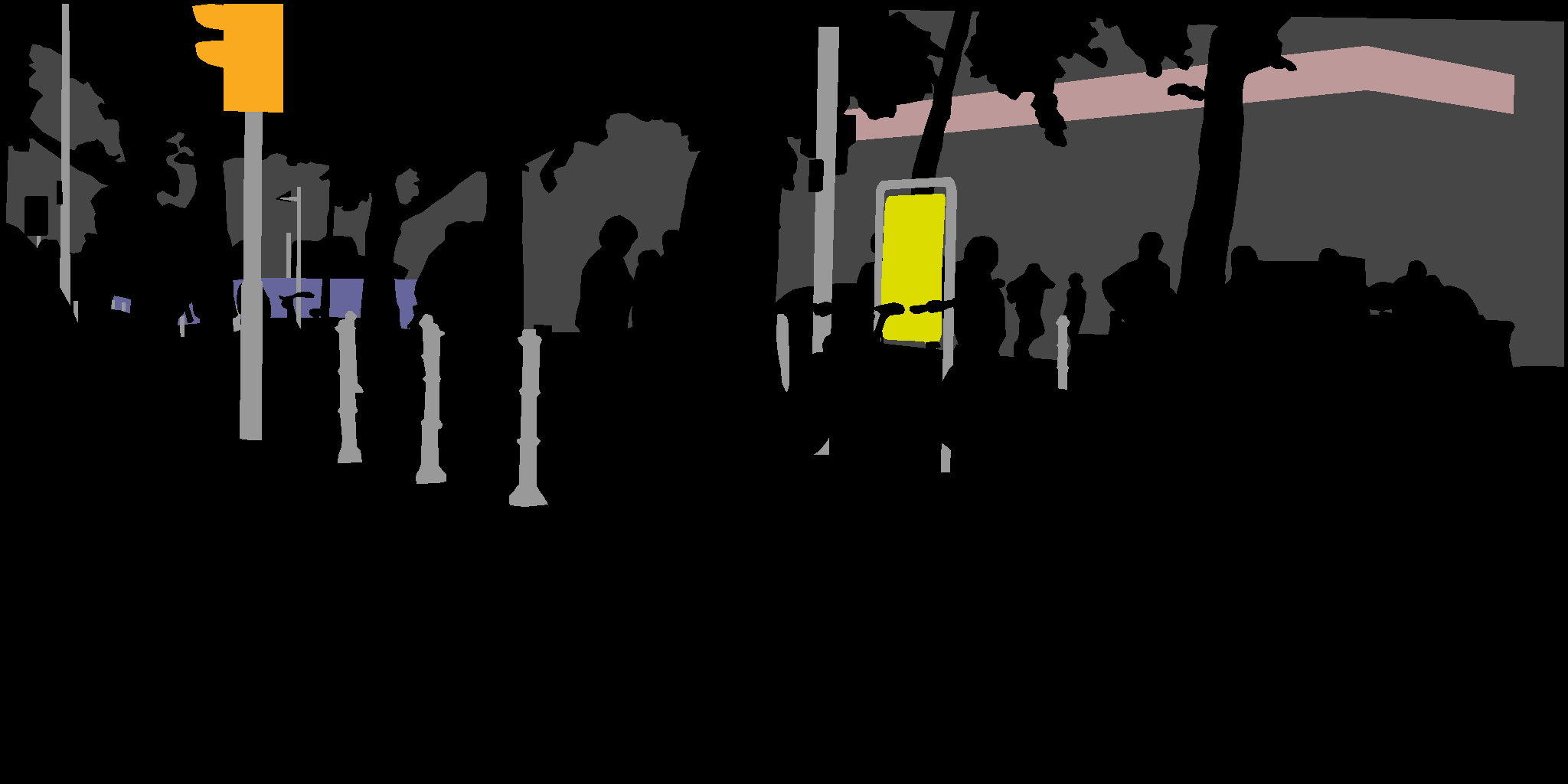} & \includegraphics[width=0.13\linewidth]{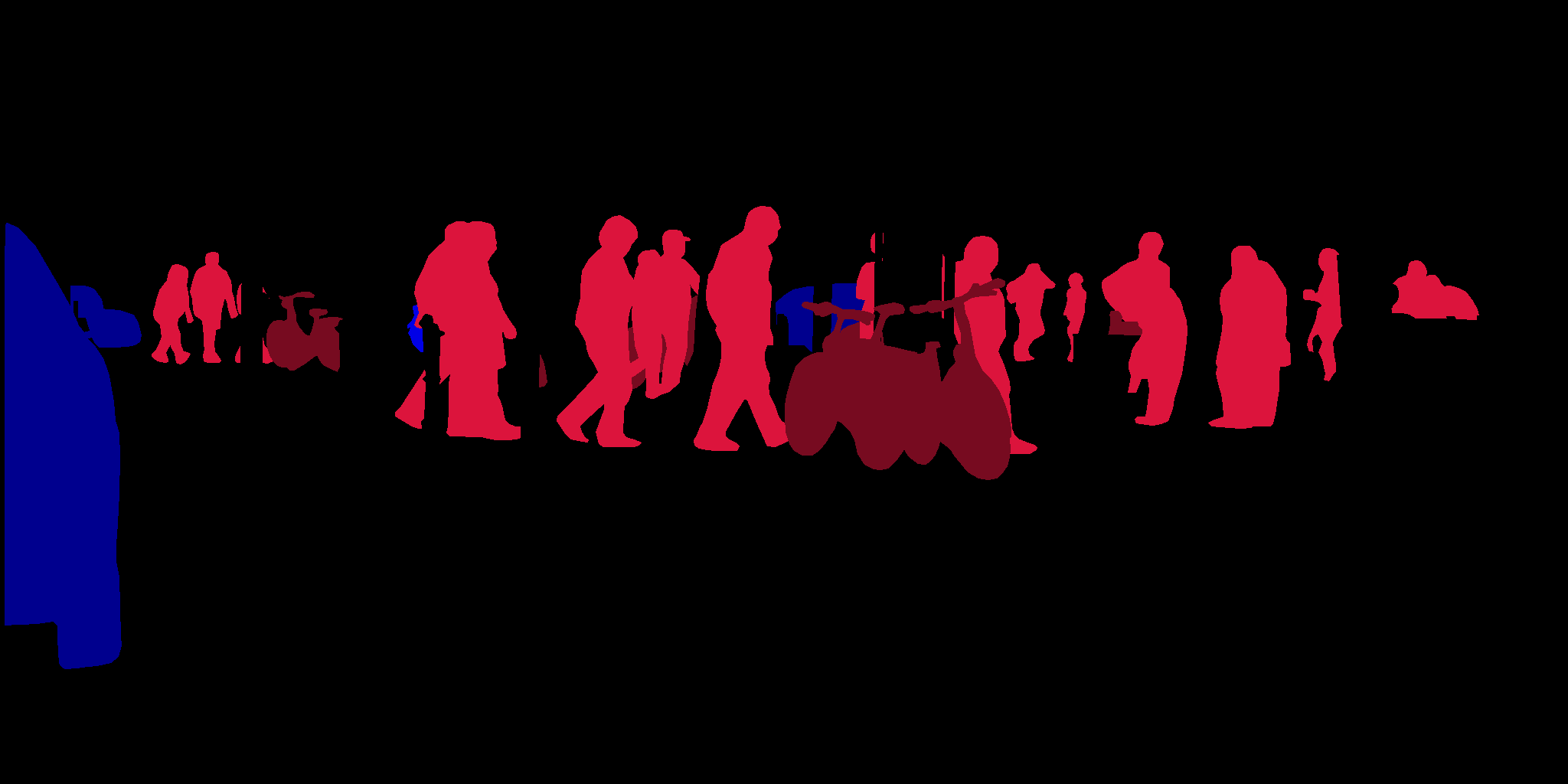} & \includegraphics[width=0.13\linewidth]{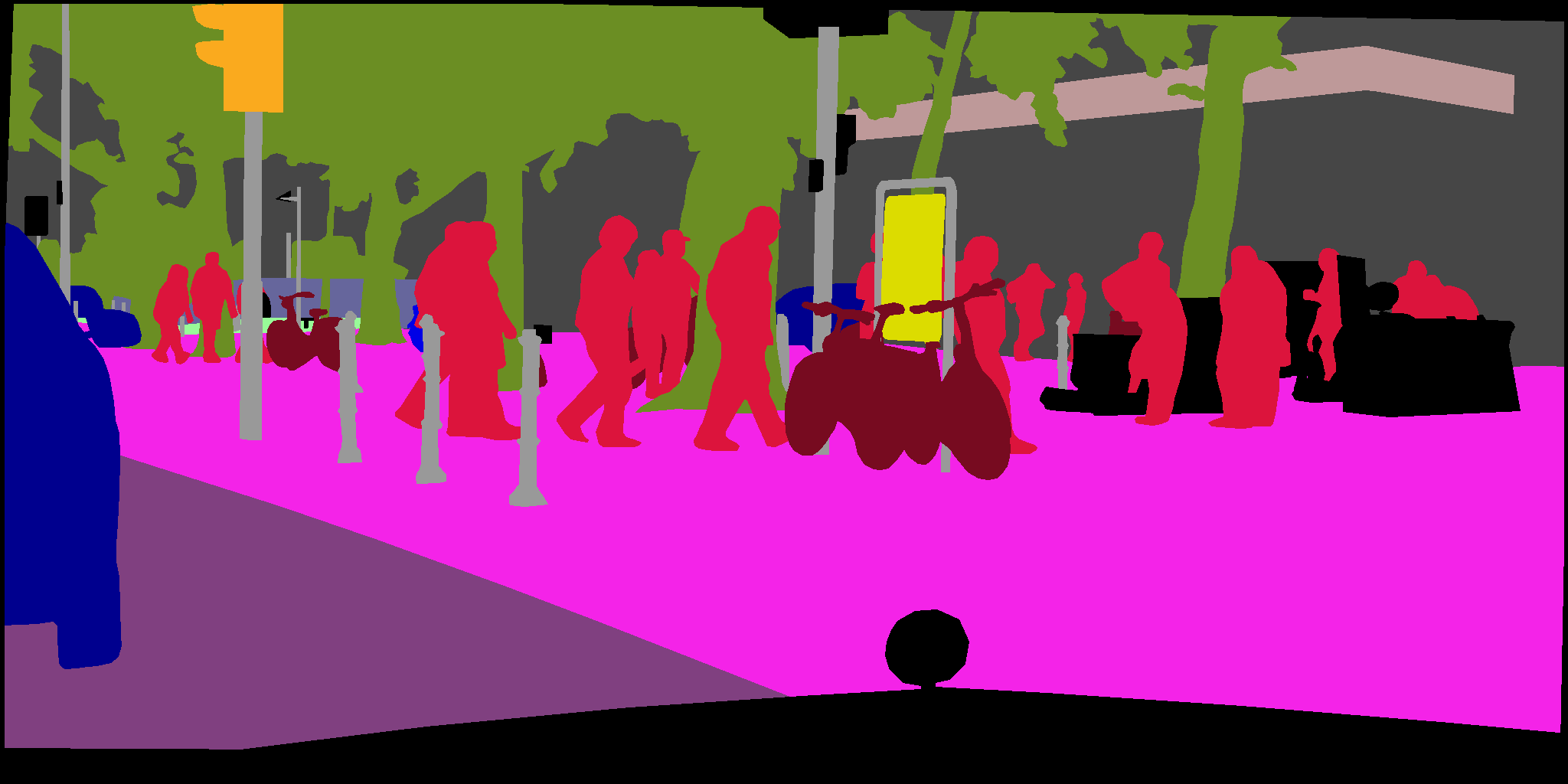}\\
    &&&&\\[-0.35cm]
    \hline
    \# images & 956 & 938 & 972 & 500\\  
  \end{tabular}  
  \label{tab:dataset_definition}
\end{table*}
In contrast to image classification, the task of semantic segmentation involves the occurrence of several classes per image at different pixel positions. Thereby, the cross-task knowledge between old and new classes can be distilled image-wise by jointly learning from the outputs of a pretrained teacher model and the labels for the new task. Previous approaches are either restricted to settings, where the additional classes have no overlap with the old ones \cite{Tasar2019}, or rely on labels for both old and new classes \cite{Michieli2019}. We, however, introduce a \textit{generally applicable technique that learns on new data solely from labels for the new classes} and outputs of a pretrained teacher model as shown in Fig.\ \ref{fig:general_overview}. Note that this setting effectively enables the extension of any pretrained semantic segmentation model to additional classes with minimum labeling effort, even allowing overlapping of labels.\par
To sum up, our contributions are as follows: Firstly, we introduce a new loss function that relies only on ground truth labels for the new classes on arbitrary data, secondly, we show how additional classes can be easily integrated in a modular fashion into arbitrary pretrained semantic segmentation models, and thirdly, we compare our new approach to various existing incremental learning methods from classification and semantic segmentation in a unified setting for the task of semantic segmentation. We demonstrate the effectiveness of the new method on the Cityscapes dataset, where our approach outperforms all other investigated incremental learning approaches for semantic segmentation. Code will be made available at \url{https://github.com/ifnspaml/CIL_Segmentation}.\par
The paper is structured as follows: In Section \ref{sec:2}, we start with an overview on related work. Afterwards, we theoretically describe our incremental learning technique for semantic segmentation in Section \ref{sec:3}, subsequently providing an experimental evaluation in Section \ref{sec:4}. Finally, we conclude the paper in Section \ref{sec:5}.

\section{Related Work}
\label{sec:2}

In this section, we give an overview to teacher-student learning with focus on semantic segmentation, as well as to approaches for incremental learning techniques in both classification and semantic segmentation.

\subsection{Semantic Segmentation}
Semantic segmentation aims at classifiying each pixel of an image. Today's state-of-the-art architectures for semantic segmentation \cite{Chen2018a, RotaBulo2018, Zhuang2018} are based upon the pioneer work by Long \etal \cite{Long2015}, where they proposed to use fully-convolutional networks (FCNs). Recently, a large interest in developing extremely efficient architectures arised \cite{Romera2018, Yu2018, Zhao2018a, Li2019a, Orsic2019}.\par
In this work, we use the \texttt{ERFNet} of Romera\ \etal\ \cite{Romera2018}, due to its simple yet effective and efficient architectural design.

\subsection{Knowledge Distillation}
Transferring the knowledge of one or an ensemble of teacher deep neural networks (DNNs) into a single student DNN is often referred to as knowledge distillation. Hinton \etal \cite{Hinton2014} showed that training the student DNN on the soft outputs of the teacher DNN is very effective. Current state-of-the-art methods try to further increase the information transfer from the teacher DNN by incorporating various intermediate feature representations and novel loss formulations during training \cite{Romero2015, Yim2017, Baer2019, Liu2019b, He2019}.\par
In this work, we use knowledge distillation combined with incremental learning to extend the set of detectable classes of a semantic segmentation DNN, while at the same time preserving its ability in detecting the original set of classes.

\subsection{Incremental Learning}

Incrementally learning several tasks with one neural network exhibits the problem of catastrophic forgetting, meaning that learning the new task is typically accompanied by forgetting the knowledge about the old task. This problem has been examined by several recent works in different settings \cite{Jung2016, Kirkpatrick2016, Rebuffi2017}. A possible solution for classification has been introduced by Li \etal \cite{Li2018c}, who propose to jointly learn from the outputs of a pretrained teacher model for the old classes and labels for the new classes. As this approach is not yet able to properly distinguish between old and new classes, later approaches \cite{Lee2019b, Castro2018} utilized memory for further training of the old classes which are selected from the dataset the model was initially trained on.\par
Previous approaches in semantic segmentation either employ a binary classification loss between each class and the background \cite{Tasar2019}, thereby assuming non-overlapping classes when extending the model, or they make use of labels for old and new classes alike \cite{Michieli2019}. In contrast, we do only use training labels for the new classes and do not assume new classes to have no overlap with the old classes, allowing the application to more general use cases.

\section{Incremental Learning for Semantic Segmentation}
\label{sec:3}

We conduct our main experiments with the classes defined by the Cityscapes dataset \cite{Cordts2016}. We divide the official Cityscapes training dataset into three training subsets of approximately equal size, as described by Table \ref{tab:dataset_definition}. The idea of our experimental setup is now that each training subset has a different subset of labeled classes, which are mutually distinct (no requirement!). One might notice that we did not use the images from the city of Erfurt, which were held back for a potential validation set, however as we train for a fixed number of epochs (no early stopping) and did not tune any hyperparameters, we did not use this set after all. In the first training stage $\mathrm{T}_1$ we train on the first subset $\mathcal{D}_1$ to learn the first set of classes $\mathcal{S}_1$ (cf. Table \ref{tab:dataset_definition}), while we use the other two subsets $\mathcal{D}_2$ and $\mathcal{D}_3$ (training stages $\mathrm{T}_2$ and $\mathrm{T}_3$, respectively) to incrementally extend the model to the second set of classes $\mathcal{S}_2$, and finally also to the third set of classes $\mathcal{S}_3$.\par
In the following, we mathematically describe the task of semantic segmentation. Afterwards, we introduce knowledge distillation for semantic segmentation, followed by several baselines for incremental learning as well as our novel class-incremental learning technique, not relying on old data or labels. For simplicity, we theoretically describe only the extension from class set $\mathcal{S}_1$ to class set $\mathcal{S}_2$, as the additional extension to class set $\mathcal{S}_3$ is straightforward by redefining the stage $\mathrm{T}_2$ segmentation model as teacher model for stage $\mathrm{T}_3$.

\subsection{Supervised Semantic Segmentation} 

In this section, we define how our semantic segmentation setup can be trained with supervision from pixel-wise class labels. It will be later used for the definition of our incremental learning methods. The setup takes an input image $\boldsymbol{x} \in \mathbb{G}^{H\times W\times C}$, where we define $\mathbb{G}$ as the set of gray values $\mathbb{G} = \left\lbrace 0, 1, ..., 255 \right\rbrace$ and $H$, $W$, and $C=3$ as the height, the width, and the number of channels of an image, respectively. Our network converts the image to output probabilities $\boldsymbol{y}\in \mathbb{I}^{H\times W \times |\mathcal{S}|}$, where $\mathbb{I} = \left[0,1\right]$ and $|\mathcal{S}|$ is the number of predicted classes. For each image pixel index $i \in \mathcal{I} = \left\lbrace 1,2,...,H\cdot W\right\rbrace$ and class $s\in \mathcal{S}$, $y_{i,s}$ can be interpreted as the posterior probability that the color pixel $\boldsymbol{x}_i\in \mathbb{G}^{C}$ belongs to class $s$. The pixel-wise elements $m_i\in \mathcal{S}$ of the segmentation mask $\boldsymbol{m}\in \mathcal{S}^{H\times W}$ are computed according to $m_i = \argmax_{s \in \mathcal{S}} y_{i, s}$, assigning a class to each pixel. The network is trained using the cross-entropy loss function

\begin{equation}
J^{\mathrm{ce}}\left(\overline{\boldsymbol{y}}, \boldsymbol{y}\right) = -\frac{1}{|\mathcal{I}_{\mathcal{S}}|}\sum_{i \in\mathcal{I}_{\mathcal{S}}}\sum_{s \in\mathcal{S}} \overline{y}_{i,s} \log\left(y_{i,s}\right)
\label{eq:crossentropy_loss}
\end{equation}
between the output probabilities $\boldsymbol{y}$ and the one-hot encoded class labels $\overline{\boldsymbol{y}}$. Note that the sum is only taken over the set of all \textit{labeled} pixels $\mathcal{I}_{\mathcal{S}} \subset \mathcal{I}$, where we only consider classes defined by $\mathcal{S}$. The training of stage $\mathrm{T}_1$ is performed purely according to (\ref{eq:crossentropy_loss}), with target class labels $\overline{\boldsymbol{y}}^{\mathcal{D}_1, \mathcal{S}_1}$ from dataset $\mathcal{D}_1$ and class set $\mathcal{S}_1$, and with predicted output probabilities $\boldsymbol{y}^{\mathrm{T}_1, \mathcal{S}_1}$ over classes $\mathcal{S}_1$ of the model trained in stage $\mathrm{T}_1$, yielding a total loss of $J^{\mathrm{ce}}\left( \overline{\boldsymbol{y}}^{\mathcal{D}_{1}, \mathcal{S}_{1}}, \boldsymbol{y}^{\mathrm{T}_{1}, \mathcal{S}_{1}}\right)$. An upper performance bound of incremental learning methods after training stage $\mathrm{T}_2$ is presented by a single-stage (\textbf{SS}) training involving all labeled classes (here: classes $\mathcal{S}_1$ and $\mathcal{S}_2$) in all used datasets at once by

\begin{equation}
	J^{\textbf{SS}} = J^{\mathrm{ce}}\left( \overline{\boldsymbol{y}}^{\mathcal{D}_{1,2}, \mathcal{S}_{1,2}}, \boldsymbol{y}^{\mathrm{T}_{1}, \mathcal{S}_{1,2}}\right).
	\label{eq:static_train}
\end{equation}
Note that in this upper bound case we use labels for all classes $\mathcal{S}_{1,2} = \mathcal{S}_1 \cup \mathcal{S}_2$ inside each image of all datasets $\mathcal{D}_{1,2} = \mathcal{D}_1 \cup \mathcal{D}_2$, thereby deviating from the definitions in Table \ref{tab:dataset_definition}. Consequently, this means that all investigated incremental learning methods (including all re-implemented baselines) have seen less labels, as they only see a subset of labeled classes per training stage.

\subsection{Knowledge Distillation} 

Since in our incremental learning approach we only use labels for the additional classes $\mathcal{S}_2$, we define how the knowledge about the old classes $\mathcal{S}_1$ can be distilled from a strong already trained teacher model. In this case one can use the output probabilities $\tilde{\boldsymbol{y}} = \boldsymbol{y}^{\mathrm{T}_1, \mathcal{S}_1}$ of this teacher model to replace the one-hot encoded labels $\overline{\boldsymbol{y}} = \overline{\boldsymbol{y}}^{\mathcal{D}_1, \mathcal{S}_1}$ in (\ref{eq:crossentropy_loss}), which is known as knowledge distillation \cite{Hinton2014}. As the teacher model supplies soft outputs for \textit{all} pixels, in this case we have $\mathcal{I}_{\mathcal{S}} = \mathcal{I}_{\mathcal{S}_1} =\mathcal{I}$, yielding the knowledge distillation loss

\begin{equation}
J^{\mathrm{kd}}\left(\tilde{\boldsymbol{y}}, \boldsymbol{y}\right) = -\frac{1}{|\mathcal{I}|}\sum_{i \in\mathcal{I}}\sum_{s \in\mathcal{S}} \tilde{y}_{i,s} \log\left(y_{i,s}\right),
\label{eq:distillation_loss}
\end{equation}
which can be applied in training stage $\mathrm{T}_2$ to preserve the knowledge about the already learned classes.

\begin{figure}[t]
	\centering	
	\includegraphics[width=1.0\linewidth]{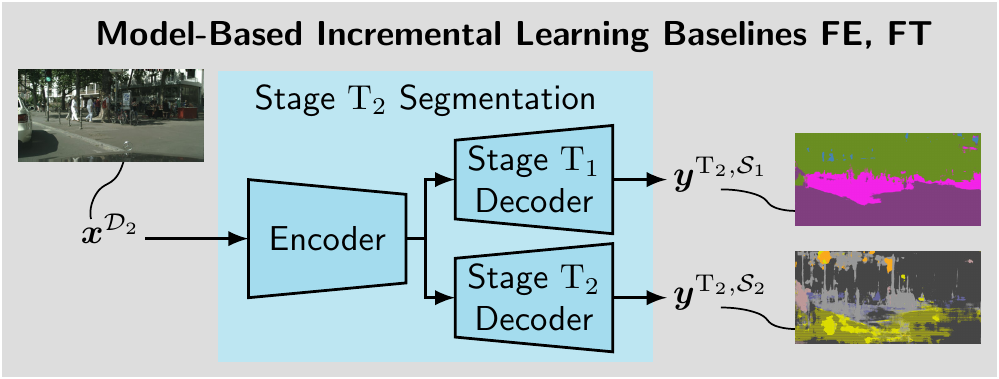}
	\caption{\textbf{Model-based incremental learning baselines} (\textbf{FE}) and (\textbf{FT}): Second training stage $\mathrm{T}_2$ segmentation model.}
	\label{fig:model_vs_data}
\end{figure} 
\begin{figure*}[t]
	\centering
	\subfloat[][Learning without forgetting (\textbf{LWOF}) baseline\label{fig:incremental_learning_a}]{\includegraphics[width=0.49\linewidth]{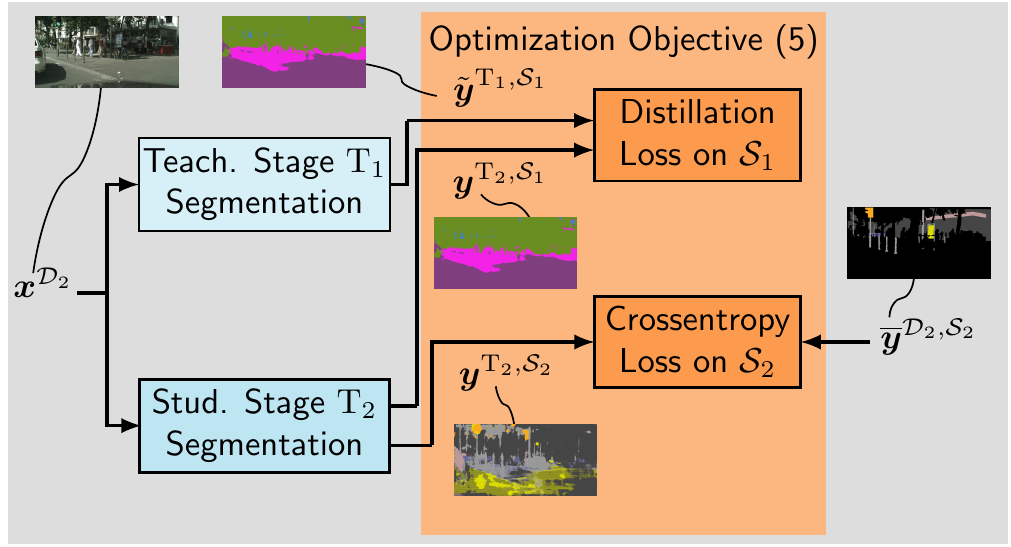}\quad}
	\subfloat[][Learning with memory (\textbf{LWM}) baseline\label{fig:incremental_in_the_wild_b}]{\includegraphics[width=0.49\linewidth]{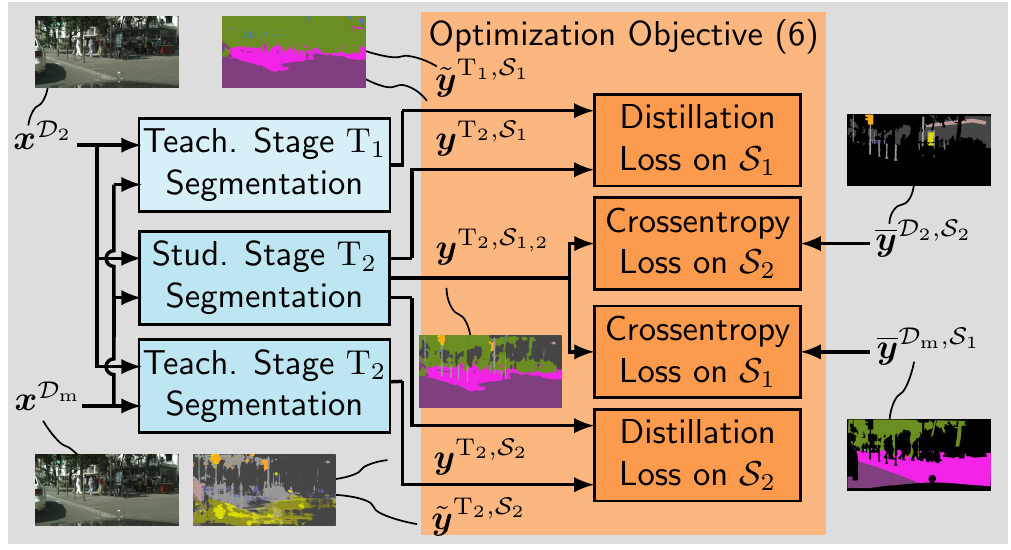}}\\[-0.08cm]
	\subfloat[][\textbf{Michieli} \cite{Michieli2019} baseline\label{fig:michieli_loss_c}]{\includegraphics[width=0.49\linewidth]{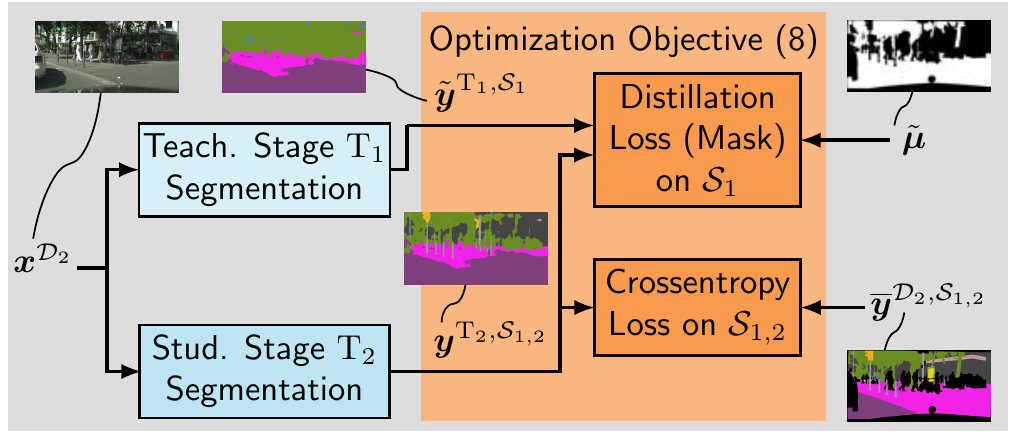}\quad}
	\subfloat[][Ours: Class-incremental learning (\textbf{CIL}) w/o old data or old labels\label{fig:coupled_distillation_d}]{\includegraphics[width=0.49\linewidth]{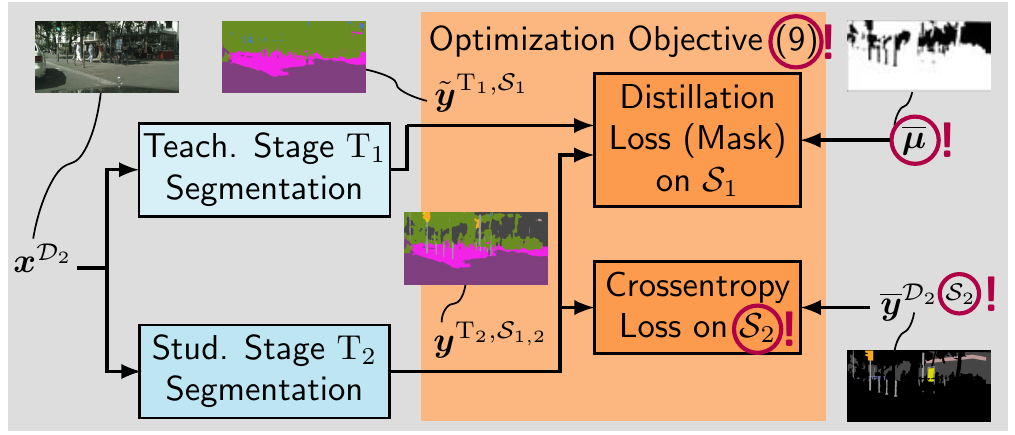}}
	\caption{\textbf{Stage} $\mathrm{T}_2$ \textbf{loss computations for different teacher-based approaches}. The former stage $\mathrm{T}_1$ segmentation becomes a teacher network now in stage $\mathrm{T}_2$. The student in stage $\mathrm{T}_2$ becomes the teacher in stage $\mathrm{T}_3$. We implement three teacher-based baseline methods, where (a) \textbf{LWOF} and (b) \textbf{LWM} are adopted from classification, and (c) is a reimplementation of \textbf{Michieli} \etal\ \cite{Michieli2019} in our framework. In (d) we present our new \textbf{CIL} approach, novelties are marked in \textcolor{tu3}{red}: Most importantly, only new data $\boldsymbol{x}^{\mathcal{D}_2}$ and labels for new classes $\mathcal{S}_2$ are needed in stage $\mathrm{T}_2$.}
	\label{fig:data_based_incremental_learning_techniques}
	\vspace{-0.1cm}
\end{figure*}

\subsection{Model-Based Baselines: Feature Extraction (\textbf{FE}) and Fine-Tuning (\textbf{FT})}

For the scope of our work, model-based incremental learning involves the training of a second set of classes $\mathcal{S}_2$ for a model that is already able to predict a first set of classes $\mathcal{S}_1$ by leaving some of the network weights fixed. Here, we compare to two baseline model-based incremental learning methods, namely feature extraction (\textbf{FE}) and fine-tuning (\textbf{FT}) \cite{Li2018c}, which extend the encoder-decoder structure of the teacher-model by a second decoder head (see Fig.\ \ref{fig:model_vs_data}). During training, the optimization is only done using the outputs of the second decoder head, providing the probability distribution $\boldsymbol{y}^{\mathrm{T}_{2}, \mathcal{S}_{2}}$ of the model in training stage $\mathrm{T}_{2}$ over the second set of classes $\mathcal{S}_{2}$. The loss functions for fine-tuning (\textbf{FT}) and feature extraction (\textbf{FE}), respectively, are defined equally as

\begin{equation}
	J^{\textbf{FE}} = J^{\textbf{FT}} = J^{\mathrm{ce}}\left(\overline{\boldsymbol{y}}^{\mathcal{D}_{2}, \mathcal{S}_{2}},  \boldsymbol{y}^{\mathrm{T}_{2}, \mathcal{S}_{2}}\right).
	\label{eq:fine_tune_feat_extract}
\end{equation}
While both methods have the same loss function, the difference is that for feature extraction (\textbf{FE}) only the second decoder head is trained, while for fine-tuning (\textbf{FT}) both the encoder and the second decoder are trained.\footnote{Usually, in classification, the splitting of the model takes place only in the last fully-connected layer \cite{Li2018c}. However, we argue that this method presents an unfair comparison for fully-convolutional networks as for those the last layer has not equally much influence. Therefore, we choose to add a full second decoder head during each extension stage as shown in Fig. \ref{fig:model_vs_data}.}

\subsection{Teacher-Based Baseline: Learning w/o Forgetting (\textbf{LWOF})}
\label{sec:LWOF}

The first implemented \textit{teacher-based} baseline is the learning without forgetting method (\textbf{LWOF}) \cite{Li2018c}, which we transfer from classification to semantic segmentation. The basic idea (cf. Fig. \ref{fig:incremental_learning_a}) is to have one output layer for all classes, where the output probabilities over the old classes $\boldsymbol{y}^{\mathrm{T}_{2}, \mathcal{S}_{1}}$ are taken and the knowledge distillation loss (\ref{eq:distillation_loss}) is computed w.r.t.~the output $\tilde{\boldsymbol{y}} = \tilde{\boldsymbol{y}}^{\mathrm{T}_{1}, \mathcal{S}_{1}}$ from the stage $\mathrm{T}_1$ teacher model of the same input image. Additionally, the output probabilities over the additional classes $\boldsymbol{y}^{\mathrm{T}_{2}, \mathcal{S}_{2}}$ are taken and the cross-entropy loss (\ref{eq:crossentropy_loss}) w.r.t.\ the labels $\overline{\boldsymbol{y}}^{\mathcal{D}_{2}, \mathcal{S}_{2}}$ is computed. The complete loss for the learning without forgetting method (\textbf{LWOF}) can thereby be written as
\begin{equation}
J^{\textbf{LWOF}} = J^{\mathrm{ce}}\left(\overline{\boldsymbol{y}}^{\mathcal{D}_{2}, \mathcal{S}_{2}}, \boldsymbol{y}^{\mathrm{T}_{2}, \mathcal{S}_{2}}\right) + J^{\mathrm{kd}}\left( \tilde{\boldsymbol{y}}^{\mathrm{T}_{1}, \mathcal{S}_{1}}, \boldsymbol{y}^{\mathrm{T}_{2}, \mathcal{S}_{1}}\right).
\label{eq:learning_without_forget}
\end{equation}

\subsection{Teacher-Based Baseline: Learning With Memory (\textbf{LWM})}
\label{sec:LWM}

The learning without forgetting method (\textbf{LWOF}) as well as the fine-tuning (\textbf{FT}) and the feature extraction (\textbf{FE}) methods have the disadvantage that they take the output probabilities over old and new classes separately, making it difficult for the network to distinguish between old and new classes. A possible solution to this problem is to introduce a memory dataset $\mathcal{D}_\mathrm{m}$ (cf. \cite{Lee2019b}), consisting of image-label pairs from the dataset the model was initially trained on. All memory samples are selected using the importance score introduced in \cite{Lee2019b}. As shown in Fig. \ref{fig:incremental_in_the_wild_b}, images from both $\mathcal{D}_\mathrm{m}$ and $\mathcal{D}_2$ are given to the networks. To balance the losses, an additional teacher model is trained solely on the new classes $\mathcal{S}_2$. Thereby, two distillation losses are computed over the class-specific probabilities of the student model w.r.t~the output probabilities of the corresponding teacher models. Specifically, in addition to the distillation loss from (\ref{eq:learning_without_forget}), an additional distillation loss between the output probabilities $\boldsymbol{y}^{\mathrm{T}_{2}, \mathcal{S}_{2}}$ of the student model and the output probabilities $\tilde{\boldsymbol{y}}^{\mathrm{T}_{2}, \mathcal{S}_{2}}$ of the teacher model for stage $\mathrm{T}_2$ is applied.\par
To enable the network to distinguish between old and new classes, the output probability $\boldsymbol{y}^{\mathrm{T}_{2}, \mathcal{S}_{1,2}}\in \mathbb{I}^{H\times W \times |\mathcal{S}_{1,2}|}$ over all classes is also computed. For images from the second training set $\mathcal{D}_2$ the cross-entropy loss between the slice $\boldsymbol{y}^{\mathrm{T}_{2}, \mathcal{S}_{1,2}}_{s\in\mathcal{S}_2}\in \mathbb{I}^{H\times W \times |\mathcal{S}_{2}|}$ of the output probabilities corresponding to the old classes and the labels $\overline{\boldsymbol{y}}^{\mathcal{D}_{2}, \mathcal{S}_{2}}$ is taken. On the other hand, for images corresponding to the memory dataset $\mathcal{D}_{\mathrm{m}}$, the cross-entropy loss between $\boldsymbol{y}^{\mathrm{T}_{2}, \mathcal{S}_{1,2}}_{s\in\mathcal{S}_1} \in \mathbb{I}^{H\times W \times |\mathcal{S}_{1}|}$ and the labels $\overline{\boldsymbol{y}}^{\mathcal{D}_{\mathrm{m}}, \mathcal{S}_{1}}$ is computed. The complete training loss can thereby be written as

\begin{align}
	J^{\textbf{LWM}} &= J^{\mathrm{ce}}\left(\overline{\boldsymbol{y}}^{\mathcal{D}_{2}, \mathcal{S}_{2}}, \boldsymbol{y}^{\mathrm{T}_{2}, \mathcal{S}_{1,2}}_{s\in\mathcal{S}_2}\right) + J^{\mathrm{ce}}\left(\overline{\boldsymbol{y}}^{\mathcal{D}_{m}, \mathcal{S}_{1}}, \boldsymbol{y}^{\mathrm{T}_{2}, \mathcal{S}_{1,2}}_{s\in\mathcal{S}_1}\right)\notag\\
	&+ J^{\mathrm{kd}}\left( \tilde{\boldsymbol{y}}^{\mathrm{T}_{1}, \mathcal{S}_{1}}, \boldsymbol{y}^{\mathrm{T}_{2}, \mathcal{S}_{1}}\right) + J^{\mathrm{kd}}\left(\tilde{\boldsymbol{y}}^{\mathrm{T}_{2}, \mathcal{S}_{2}}, \boldsymbol{y}^{\mathrm{T}_{2}, \mathcal{S}_{2}}\right)
\end{align}

\subsection{Teacher-Based Baseline: \textbf{Michieli}}
\label{sec:Michieli}

Although learning with memory (\textbf{LWM}) introduces a lot of complexity, it shows that the cross-task decision between old and new classes can be learned by applying the losses on a joint output probability for all classes. Further developing this idea, Michieli \etal\ \cite{Michieli2019} introduce an incremental learning technique, where in stage $\mathrm{T}_2$ the cross-entropy is taken over all classes with labels $\overline{\boldsymbol{y}} = \overline{\boldsymbol{y}}^{\mathcal{D}_2, \mathcal{S}_{1,2}}$ for all classes, while the distillation loss is taken only at positions with labels for the old classes. As a prerequisite, we define the so-called masked knowledge distillation loss

\begin{equation}
J^{\mathrm{mkd}}\left(\tilde{\boldsymbol{y}}, \boldsymbol{y}, \boldsymbol{\mu}\right) = -\frac{1}{|\mathcal{I}_{\mu}|}\sum_{i \in\mathcal{I}}\sum_{s \in\mathcal{S}} \alpha_i \mu_i \tilde{y}_{i,s} \log\left(y_{i,s}\right),
\label{eq:distillation_loss_masked}
\end{equation}
where $\mu_i$ and $\alpha_i$ are the pixel-wise elements of a binary mask $\boldsymbol{\mu}\in \left\lbrace 0, 1 \right\rbrace^{H\times W}$ and a weight mask $\boldsymbol{\alpha}\in \mathbb{R}^{H\times W}$, respectively. The set $\mathcal{I}_{\mu}\subset \mathcal{I}$ is the set of all pixels contributing to the loss, meaning a mask value $\mu_i=1$ at pixel index $i$.\par
For our reimplementation of \cite{Michieli2019} we define the mask $\boldsymbol{\mu} = \tilde{\boldsymbol{\mu}} \in \left\lbrace 0, 1 \right\rbrace^{H\times W}$ with its pixel-wise elements $\tilde{\mu}_i$ containing a $1$ if the pixel is labeled with an old class and a $0$ otherwise. The elements $\alpha_i$ are all set to 1, meaning no individual weighting of the pixels against each other is made. Note that one needs to have labels for the old classes in order to calculate the mask $\tilde{\boldsymbol{\mu}}$. The complete training loss for our reimplementation of \cite{Michieli2019} is then given by

\begin{equation}
	J^{\textbf{Mi}} = J^{\mathrm{ce}}\left(\overline{\boldsymbol{y}}^{\mathcal{D}_{2}, \mathcal{S}_{1,2}}, \boldsymbol{y}_{}^{\mathrm{T}_{2}, \mathcal{S}_{1,2}}\right) + J^{\mathrm{mkd}}\left( \tilde{\boldsymbol{y}}^{\mathrm{T}_{1}, \mathcal{S}_{1}}, \boldsymbol{y}_{s\in\mathcal{S}_1}^{\mathrm{T}_{2}, \mathcal{S}_{1,2}}, \tilde{\boldsymbol{\mu}}\right).
\end{equation}

\subsection{Ours: Class-Incremental Learning (\textbf{CIL}) w/o Old Data or Old Labels}
Although the basic idea to take a joint output probability distribution is a convincing concept, we identify two disadvantages in Michieli's approach \cite{Michieli2019}: Firstly, Michieli's stage $\mathrm{T}_2$ training approach relies on labels for both old and new classes, and secondly, it is biased towards the old classes $\mathcal{S}_1$, which appear in both loss terms. Conclusively, we propose to apply the cross-entropy loss \textit{only over the new class pixels}. Furthermore, we apply the masked distillation loss \textit{in all other parts of the image} yielding a complete loss of

\begin{equation}
	J^{\textbf{CIL}} = J^{\mathrm{ce}}\left(\overline{\boldsymbol{y}}^{\mathcal{D}_{2}, \mathcal{S}_{2}}, \boldsymbol{y}_{s\in\mathcal{S}_2}^{\mathrm{T}_{2}, \mathcal{S}_{1,2}}\right) + J^{\mathrm{mkd}}\left( \tilde{\boldsymbol{y}}^{\mathrm{T}_{1}, \mathcal{S}_{1}}, \boldsymbol{y}_{s\in\mathcal{S}_1}^{\mathrm{T}_{2}, \mathcal{S}_{1,2}}, \overline{\boldsymbol{\mu}}\right),
	\label{eq:coupled_distill}
\end{equation}
where the mask $\overline{\boldsymbol{\mu}}$ with its elements $\overline{\mu}_i$ contains a $0$ if the pixel is labeled with a new class $s\in \mathcal{S}_2$ and a $1$ otherwise. Thereby, we do not re-use any labels for the old classes and can very efficiently include new classes into a pretrained semantic segmentation model.\par
Moreover, we noticed that soft labels as used in the distillation losses (\ref{eq:distillation_loss}) and (\ref{eq:distillation_loss_masked}) yield only small gradients compared to one-hot encoded labels. Furthermore, we also noticed that classes which do only appear in small regions of the image (e.g., traffic signs or traffic lights) tend to have a distribution being rather soft. Hence, these classes are usually not well-represented by the losses (\ref{eq:distillation_loss}) and (\ref{eq:distillation_loss_masked}), yielding an unusually low performance on these classes. Desiring to emphasize this class-wise differing uncertainty of the teacher model and intending to put more weight to the gradients of these difficult yet important classes, we define the pixel-wise weights $\alpha_i$ in (\ref{eq:distillation_loss_masked}) entropy-based, such that the more uniform the soft label distribution is, the bigger the weight. Conclusively, we define
\begin{equation}
	\alpha_i = 1 + \left(-\sum_{s\in\mathcal{S}} \tilde{y}_{i,s} \cdot \log_2\left(\tilde{y}_{i,s}\right) \right).
	\label{eq:entropy_weights}
\end{equation}
Note that these weights are applied only in the masked knowledge distillation loss term $J^{\mathrm{mkd}}$ of our approach described by (\ref{eq:coupled_distill}).

\begin{table*}[t]
  \centering
  \caption{Comparison between \textbf{model-based methods and our method} on the Cityscapes validation dataset, which is defined as our test set (cf. Table \ref{tab:dataset_definition}). \textbf{Best results} are marked in \textbf{boldface}, \underline{second best} results are \underline{underlined}. The single-stage training method (\textbf{SS}) as an upper performance bound is out of competition. mIoU values in [\%].}
  \begin{tabular}{l|l|ccccc}
    \textbf{Method} & Training Stages & $\mathrm{mIoU}_{\mathrm{Task\; 1}}$ & $\mathrm{mIoU}_{\mathrm{Task\; 2}}$ & $\mathrm{mIoU}_{\mathrm{Task\; 1 \cup 2}}$ & $\mathrm{mIoU}_{\mathrm{Task\; 3}}$ & $\mathrm{mIoU}_{\mathrm{Task\; 1\cup 2 \cup 3}}$  \\
    \hline\hline
    \multicolumn{2}{c|}{\textbf{SS} Training with classes $\mathcal{S}_{1,2}$} & $87.7$ & $66.1$ & $70.7$ & - & -\\
    \hline
    \textbf{FT} & $\mathrm{T}_1$, $\mathrm{T}_2$ & $37.0$ & $\underline{59.7}$  & $29.2$ & - & -\\
    \textbf{FE} & $\mathrm{T}_1$, $\mathrm{T}_2$ & $\textbf{86.2}$& $45.4$ & $\underline{44.8}$ & - & -\\
    \textbf{CIL} (ours) & $\mathrm{T}_1$, $\mathrm{T}_2$ & $\underline{85.5}$& $\textbf{63.9}$ & $\textbf{67.7}$ & - & -\\
    \hline\hline
    \multicolumn{2}{c|}{\textbf{SS} Training with classes $\mathcal{S}_{1,2,3}$} & $88.1$ & $65.6$ & $70.8$ & $63.8$ & $64.4$\\
    \hline
    \textbf{FT} & $\mathrm{T}_1$, $\mathrm{T}_2$, $\mathrm{T}_3$ & $24.8$& $18.0$ & $10.8$ & $\underline{56.5}$ & $10.4$\\
    \textbf{FE} & $\mathrm{T}_1$, $\mathrm{T}_2$, $\mathrm{T}_3$ & $\textbf{86.2}$ & $\underline{45.4}$ & $\underline{44.8}$ & $37.4$ & $\underline{30.3}$\\
    \textbf{CIL} (ours) & $\mathrm{T}_1$, $\mathrm{T}_2$, $\mathrm{T}_3$ & $\underline{84.5}$ & $\textbf{57.8}$ & $\textbf{64.0}$ & $\textbf{68.4}$ & $\textbf{62.2}$\\
  \end{tabular}  
  \label{tab:comparison_model_based_data}
\end{table*}

\begin{table*}[t]
  \centering
   \caption{Comparison of different \textbf{teacher-based methods} on the Cityscapes validation set (cf. Table \ref{tab:dataset_definition}). The tag ``w/o weights'' means that we did not use the entropy-based weights from (\ref{eq:entropy_weights}). \textbf{Best results} are marked in \textbf{boldface}, \underline{second best} results are \underline{underlined}. mIoU values in [\%].}
  \begin{tabular}{l|l|ccccc}
    \textbf{Method} & Training Stages & $\mathrm{mIoU}_{\mathrm{Task\; 1}}$ & $\mathrm{mIoU}_{\mathrm{Task\; 2}}$ & $\mathrm{mIoU}_{\mathrm{Task\; 1 \cup 2}}$  & $\mathrm{mIoU}_{\mathrm{Task\; 3}}$ & $\mathrm{mIoU}_{\mathrm{Task\; 1\cup 2 \cup 3}}$  \\
    \hline\hline
    \textbf{LWOF} & $\mathrm{T}_1$, $\mathrm{T}_2$ & $85.4$ & $62.8$ & $57.2$ & - & -\\
    \textbf{LWM} & $\mathrm{T}_1$, $\mathrm{T}_2$ & $\underline{85.9}$& $61.4$ & $62.5$ & - & -\\
    \textbf{Michieli} \cite{Michieli2019} (reimplemented) & $\mathrm{T}_1$, $\mathrm{T}_2$ & $\textbf{87.0}$ & $59.7$ & $66.4$ & - & -\\
    \textbf{CIL} (ours, w/o weights) & $\mathrm{T}_1$, $\mathrm{T}_2$ & $85.5$ & $\textbf{65.8}$ & $\textbf{68.8}$ & - & -\\
    \textbf{CIL} (ours) & $\mathrm{T}_1$, $\mathrm{T}_2$ & $85.5$& $\underline{63.9}$ & $\underline{67.7}$ & - & -\\
    \hline\hline
    \textbf{LWOF} & $\mathrm{T}_1$, $\mathrm{T}_2$, $\mathrm{T}_3$ & $84.0$ & $52.6$ & $48.8$ & $64.5$ & $43.5$\\
    \textbf{LWM} & $\mathrm{T}_1$, $\mathrm{T}_2$, $\mathrm{T}_3$ & $83.9$& $57.4$ & $61.1$ & $62.0$ & $52.6$\\
    \textbf{Michieli} \cite{Michieli2019} (reimplemented) & $\mathrm{T}_1$, $\mathrm{T}_2$, $\mathrm{T}_3$ & $\textbf{86.5}$ & $\textbf{60.5}$ & $\textbf{66.6}$ & $54.5$ & $58.7$\\
    \textbf{CIL} (ours, w/o weights) & $\mathrm{T}_1$, $\mathrm{T}_2$, $\mathrm{T}_3$ & $83.9$ & $52.7$ & $61.1$ & $\underline{67.9}$ & $\underline{59.5}$\\
    \textbf{CIL} (ours) & $\mathrm{T}_1$, $\mathrm{T}_2$, $\mathrm{T}_3$ & $\underline{84.5}$ & $\underline{57.8}$ & $\underline{64.0}$ & $\textbf{68.4}$ & $\textbf{62.2}$
  \end{tabular}  
  \label{tab:comparison_data_based}
\end{table*}

\section{Experiments and Discussion}
\label{sec:4}

In this section, we start with describing our experimental setup. Afterwards, we first compare our method to model-based methods and then to various teacher-based methods on the Cityscapes dataset.

\subsection{Experimental Setup}

We conduct our experiments utilizing the deep learning library \texttt{PyTorch} \cite{Paszke2019}. The network architecture is the \texttt{ERFNet} encoder-decoder structure \cite{Romera2018}. In our experiments we keep the network structure unchanged except for the last layer, whose number of feature maps corresponds to the number of predicted classes. For model-based training (\textbf{FE}, \textbf{FT}) we also introduce a second network head, which has the same structure as the first one.\par
In contrast to the original \texttt{ERFNet} model, we apply additional augmentation techniques: random rotation of ($\pm 2^\circ$), random rescaling with factors $\left[1.0, 1.5\right]$, random cropping to $640\times 192$, random brightness ($\pm 0.2$), contrast ($\pm 0.2$), saturation ($\pm 0.2$) and hue ($\pm 0.1$). While this slightly decreases our performance on the Cityscapes dataset, it enables testing of our models on more general data domains, which previously was not possible due to the high overfitting of the original model to the Cityscapes dataset.\par
We train all models for 200 epochs utilizing the Adam \cite{Kingma2015} optimizer with an initial learning rate of $5\cdot 10^{-4}$, a batch size of $6$ and weight decay of $3\cdot 10^{-4}$ due to the small size of the training sets. We use a polynomial learning rate scheduling as in \cite{Zhao2016a} with a power of $0.9$.\par
We start by training a segmentation model in stage $\mathrm{T}_1$ on subset $\mathcal{D}_1$ for classes $\mathcal{S}_1$, while afterwards extending our model in stages $\mathrm{T}_2$ and $\mathrm{T}_3$ by the classes $\mathcal{S}_2$ and $\mathcal{S}_3$, utilizing the subsets $\mathcal{D}_2$ and $\mathcal{D}_3$, respectively. For all \textit{model-based} methods, we initialize the stage $\mathrm{T}_2$ segmentation model with the pre-trained weights from the first training stage $\mathrm{T}_1$, except for the stage $\mathrm{T}_2$ decoder (cf. Fig. \ref{fig:model_vs_data}), which is initialized randomly. In contrast, the initialization of all \textit{teacher-based} models neither relies on the network architecture nor on the pre-trained weights of the stage $\mathrm{T}_1$ segmentation model\footnote{In fact, one can initialize the weights randomly at each training stage, which we adopted due to improved results, and even one could use a completely different network structure in stage $\mathrm{T}_2$ which could be used to upscale or downscale an existing model according to hardware requirements. However, for comparability of the different methods we keep the initial network architecture unchanged and only extend the last layer by additional output feature maps for the additional classes.}, as the whole training process is based on loss functions utilizing labels or outputs of a teacher network.\par
We evaluate the experiments using the mean intersection over union (mIoU) metric, see Tables \ref{tab:comparison_model_based_data}, \ref{tab:comparison_data_based}. After training stage $\mathrm{T}_2$, we compute the single-task metrics $\mathrm{mIoU}_{\mathrm{Task\; 1}}$ and $\mathrm{mIoU}_{\mathrm{Task\; 2}}$ over the classes $\mathcal{S}_1$ and $\mathcal{S}_2$, respectively, as well as for the first time the cross-task metric $\mathrm{mIoU}_{\mathrm{Task\; 1 \cup 2}}$ over classes $\mathcal{S}_{1,2}$. After training stage $\mathrm{T}_3$, we additionally compute the single-task metric $\mathrm{mIoU}_{\mathrm{Task\; 3}}$ and the cross-task metric $\mathrm{mIoU}_{\mathrm{Task\; 1\cup 2 \cup 3}}$. Note that the single-task metrics only measure the capability of a model to distinguish between classes from a single class subset belonging to the respective task. However, when extending a segmentation model to additional classes it is also important that a model can distinguish between classes from several subsets, which are learned incrementally. Conclusively, the cross-task metrics present a more meaningful quality indicator as they show the model's capability in distinguishing classes from several subsets. Also note that for all metrics we compute the output probabilities of a model only over the feature maps of the classes considered for the respective mIoU score.
\begin{figure*}[t]
	\centering	
	\includegraphics[width=1.0\linewidth]{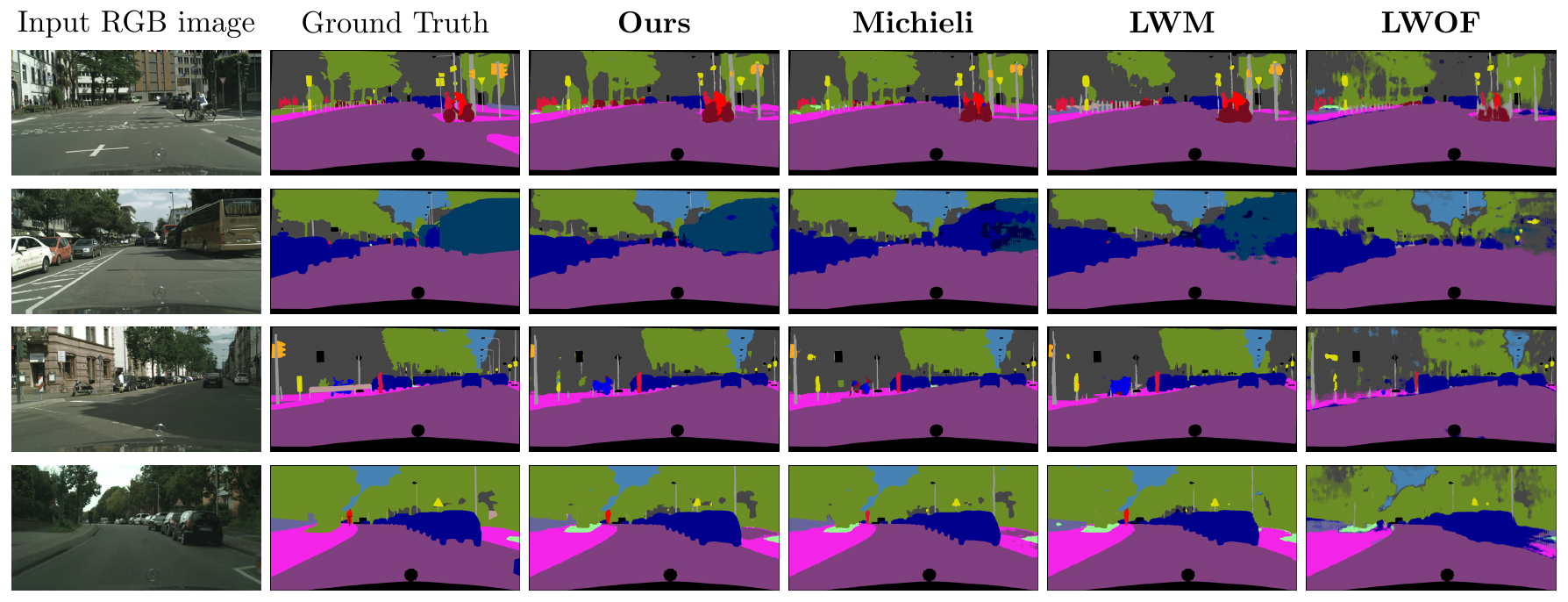}
	\caption{Qualitative \textbf{examples of teacher-based methods} after training stage $\mathrm{T}_3$ in comparison to ground truth labels. The figure is best viewed on screen and in color.}
	\label{fig:qualitative_results}
\end{figure*} 
\subsection{Model-Based Methods vs. Teacher-Based Methods}

To be able to determine how good a method actually is, we first train a model using all labels from all training subsets (cf. Table \ref{tab:dataset_definition}) as described by (\ref{eq:static_train}). This single stage (\textbf{SS}) training is out of competition for incremental learning methods as it relies on \textit{all} 19 classes to be labeled in \textit{each} image. Furthermore, the method relies solely on labeled data, while for incremental learning methods one can make use of the additional knowledge of any pretrained teacher model. However, in our case, \textbf{SS} training gives a good upper bound for the possible performance of any incremental learning method. An overview of the results of the \textbf{SS} training method and our method in comparison to model-based methods is given in Table \ref{tab:comparison_model_based_data}.\par
Our examined model-based approaches are fine-tuning (\textbf{FT}) and feature extraction (\textbf{FE}), which are described by (\ref{eq:fine_tune_feat_extract}). We observe that the \textbf{FT} approach yields a decent performance on the new task with $\mathrm{mIoU}_{\mathrm{Task\; 2}} = 59.7\%$, while the performance of the old task significantly decreases to $\mathrm{mIoU}_{\mathrm{Task\; 1}} = 37.0\%$, which is expected as we retrain the encoder without adapting the stage $\mathrm{T}_1$ decoder for the first task. Feature extraction (\textbf{FE}) solves this problem to some extent by keeping the encoder weights fixed, yielding a significantly higher performance on the first task with $\mathrm{mIoU}_{\mathrm{Task\; 1}} = 86.2\%$. However, the performance on the second task is significantly lower with $\mathrm{mIoU}_{\mathrm{Task\; 2}} = 45.4\%$, as the encoder has never been trained to extract features that are also beneficial for the second task. Both model-based approaches yield poor cross-task performance on the $\mathrm{mIoU}_{\mathrm{Task\; 1 \cup 2}}$ metric which is significantly lower than the single-stage training performance limit. This is mainly due to the fact that the two tasks are trained separately, giving no guidance on how to distinguish between classes of different tasks.\par
As can be seen, our approach significantly outperforms both model-based approaches in all metrics, except for the single-task performance on the first task, where it closely ranks second. However, on the most important cross-task metrics $\mathrm{mIoU}_{\mathrm{Task\; 1 \cup 2}}$ and $\mathrm{mIoU}_{\mathrm{Task\; 1\cup 2 \cup 3}}$, our approach outperforms both model-based approaches by a large margin, indicating that one should prefer teacher-based approaches for incremental learning of semantic segmentation, which is in accordance with the results from \cite{Li2018c} and \cite{Michieli2019}. Surprisingly, after two extensions of our model, first to the classes $\mathcal{S}_2$ and afterwards to the classes $\mathcal{S}_3$, \textit{our new} \textbf{CIL} \textit{approach reaches an $\mathrm{mIoU}_{\mathrm{Task\; 1\cup 2 \cup 3}} = 62.2\%$, which is only $2.2\%$ absolute below the upper performance bound (single-stage training), although our approach utilizes significantly less labeled classes/pixels in each image, and follows a strict incremental training protocol}.

\subsection{Comparison of Different Teacher-Based Methods}

Having identified that teacher-based methods outperform model-based methods, it is essential to know which strategy one should apply when extending a neural network for semantic segmentation. For this, we implemented two approaches from image classification \cite{Li2018c, Lee2019b}, adopting them to semantic segmentation (see Sections \ref{sec:LWOF}, \textbf{LWOF}, and \ref{sec:LWM}, \textbf{LWM}) and also one approach from semantic segmentation \cite{Michieli2019} in our setting (Section \ref{sec:Michieli}, \textbf{Michieli}). An overview over the results of all methods is given in Table \ref{tab:comparison_data_based} and some exemplary network outputs for all methods after training stage $\mathrm{T}_3$ are given in Fig. \ref{fig:qualitative_results}.\par
The learning without forgetting method (\textbf{LWOF}) achieves decent results of $\mathrm{mIoU}_{\mathrm{Task\; 1}} = 85.4\%$ and $\mathrm{mIoU}_{\mathrm{Task\; 2}} = 62.8\%$ on the single tasks. However, we observe that the overall performance $\mathrm{mIoU}_{\mathrm{Task\; 1 \cup 2}} = 57.2\%$ is significantly lower due to the fact that the two tasks are trained on two different losses giving only poor signals for the cross-task decision. The learning with memory (\textbf{LWM}) method solves this problem to some extent which can be seen by the fact that the single-task performances are similar to \textbf{LWOF}, however, the cross-task performance is significantly better with $\mathrm{mIoU}_{\mathrm{Task\; 1 \cup 2}} = 61.7\%$.\par
The \textbf{Michieli} approach \cite{Michieli2019} only optimizes the joint probability over old and new classes, which seems to be beneficial as the cross-task decision can thereby be easily learned. However, it requires labels for all classes, which makes the method rather comparable to the \textbf{SS} training method. An overall performance of $\mathrm{mIoU}_{\mathrm{Task\; 1 \cup 2}}=66.4\%$ is achieved, which is still worse than the \textbf{SS} training method ($70.7\%$). Also, \textbf{Michieli} seems to be biased towards the old classes, which occur in both loss terms, yielding the best single-task performance of all teacher-based approaches on the first task ($\mathrm{mIoU}_{\mathrm{Task\; 1}} = 87.0\%$) while achieving the worst performance on the second task ($\mathrm{mIoU}_{\mathrm{Task\; 2}} = 59.7\%$). Nevertheless, the overall performance is better than \textbf{LWOF} and \textbf{LWM}, which, however, is not surprising as \textbf{Michieli} utilizes labels for all classes.\par
Our new \textbf{CIL} approach picks up the idea of optimizing the joint probability distribution over all classes and balances the training objective, which can be seen in the higher single-task performance of $\mathrm{mIoU}_{\mathrm{Task\; 2}} = 63.9\%$, while the performance on the first task remains constantly high. In the important cross-task metrics \textit{we outperform the} \textbf{Michieli} \textit{approach} \cite{Michieli2019}, \textit{although using only labels for the new classes}. Particularly our \textbf{CIL} (w/o weights) approach improves $2.4\%$ absolute in terms of $\mathrm{mIoU}_{\mathrm{Task\; 1\cup 2}}$ upon the \textbf{Michieli} approach ($68.8\%$ vs.~$66.4\%$). Even more important, after two extension steps (all Cityscapes classes being used) our \textbf{CIL} method outperforms the \textbf{Michieli} approach by $3.5\%$ absolute in terms of $\mathrm{mIoU}_{\mathrm{Task\; 1\cup 2 \cup 3}}$ ($62.2\%$ vs.~$58.7\%$). We also outperform all other teacher-based baselines in the cross-task metrics.\par
Investigating the impact of the pixel weights from (\ref{eq:entropy_weights}), we can see that their application results in a slightly worse performance during the second stage $\mathrm{T}_2$, however, they significantly improve the performance in the third stage from $\mathrm{mIoU}_{\mathrm{Task\; 1\cup 2 \cup 3}} = 59.5\%$ to $\mathrm{mIoU}_{\mathrm{Task\; 1\cup 2 \cup 3}} = 62.2\%$, which is why we chose this model to be our final one, as during development one might need to extend a model several times.

\section{Conclusion}
\label{sec:5}

In this paper, we give an extensive comparison of incremental learning techniques for semantic segmentation. We identify weaknesses in current approaches and as a result propose a novel loss, which in contrast to past works neither relies on old labels nor on old data. We show the effectiveness of our method on the Cityscapes dataset, where we outperform all other methods in terms of mIoU by $2.4\%$ absolute (second stage) and $3.5\%$ absolute (third stage). The latter result is just $2.2\%$ absolute below the upper performance limit of single-stage training, relying on all data and labels at once. Our proposed method enables future semantic segmentation algorithms to be easily extended to new classes without relying on the dataset used for training of the initial model. Code will be made available at \url{https://github.com/ifnspaml/CIL_Segmentation}.

\bibliographystyle{IEEEtran}
\bibliography{IEEEabrv,./ifn_spaml_bibliography}

\end{document}